\documentclass[conference]{IEEEtran}
\def\IEEEtitletopspace{18pt}
\IEEEoverridecommandlockouts
\usepackage[letterpaper, top=1.91cm, bottom=1.91cm, left=1.91cm, right=1.91cm]{geometry}
\usepackage{cite}
\usepackage{amsmath,amssymb,amsfonts}
\usepackage{algorithmic}
\usepackage{graphicx}
\usepackage{textcomp}
\usepackage{xcolor}
\usepackage{hyperref}
\usepackage{enumitem}
\usepackage{makecell}
\usepackage{array}
\usepackage[bottom,flushmargin]{footmisc}
\usepackage[font=small]{caption}
\def\BibTeX{{\rm B\kern-.05em{\sc i\kern-.025em b}\kern-.08em
    T\kern-.1667em\lower.7ex\hbox{E}\kern-.125emX}}
\begin{document}


\title{Design of Q8bot: A Miniature, Low-Cost, Dynamic Quadruped Built with Zero Wires\\}
\author{Yufeng Wu\textsuperscript{1} and Dennis Hong\textsuperscript{1}}
\maketitle

\begingroup
\renewcommand\thefootnote{1}
\footnotetext{All authors are with the Department of Mechanical and Aerospace Engineering, University of California, Los Angeles, CA, USA 90095 \texttt{\{ericyufengwu, dennishong\}@g.ucla.edu}.}
\endgroup

\begin{abstract}
This paper introduces Q8bot\textsuperscript{2}, an open-source, miniature quadruped designed for robotics research and education. We present the robot's novel, zero-wire design methodology, which leads to its superior form factor, robustness, replicability, and high performance. With the size and weight similar to a modern smartphone, this standalone robot can walk hour-long on a single battery charge and can survive meter-high drops with simple repairs. Its \$300 bill of materials contains minimal off-the-shelf components, readily available custom electronics from online vendors, and structural parts that can be manufactured on hobbyist 3D-printers. A preliminary user assembly study confirms that Q8bot can be easily replicated, with an average assembly time of under one hour by a single person. With heuristic open-loop control, Q8bot is capable of a stable walking speed of 5.4 body lengths per second and a turning speed of 5 radians per second, along with other dynamic movements such as jumping and climbing moderate slopes.
\end{abstract}

\begingroup
\renewcommand\thefootnote{2}
\footnotetext{Q8bot Repository: https://github.com/EricYufengWu/q8bot}
\endgroup


\section{Introduction}
There have been significant developments in quadruped robots due to their ability to navigate complex terrains and perform versatile tasks in various applications \cite{ANYmal}, \cite{MiniCheetah}, \cite{ALPHRED}, \cite{SCALER}. Recent democratization of proprioceptive actuator design \cite{katz2018low}, actuator-level field-oriented control \cite{mevey2009sensorless}, and robot-level control methods \cite{CMPC} has made the technologies required to construct dynamic quadruped robots more accessible. While fully commercial quadruped systems are now available \cite{unitree_go2}, research communities often favor open-source hardware platforms due to their high level of accessibility and customizability \cite{Solo}, \cite{Doggo}, \cite{MEVIUS}, \cite{Pupper}.

\begin{figure}[t]  
    \centering
    \includegraphics[width=0.48\textwidth]{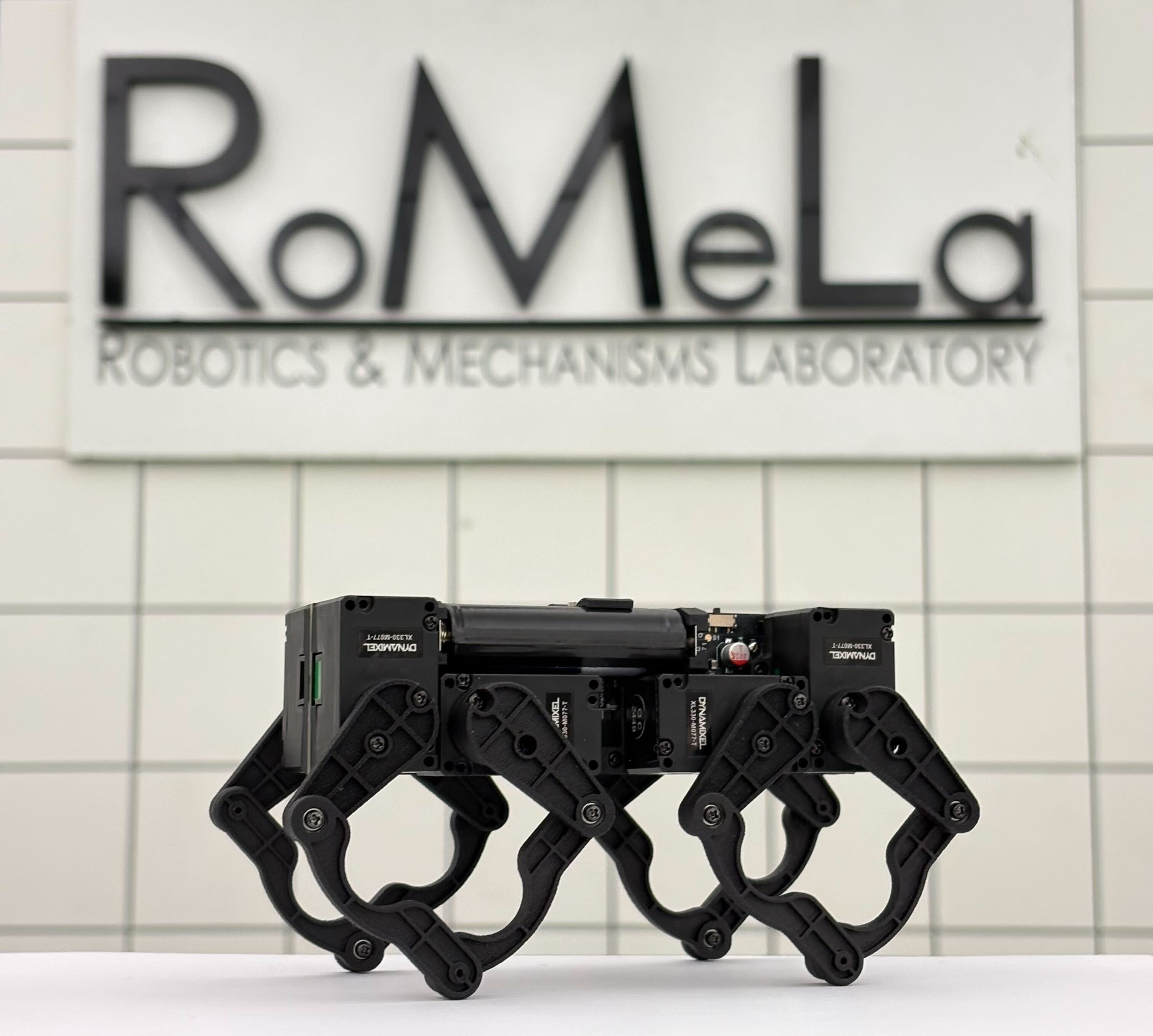}  
    \captionsetup{belowskip=-10pt}
    \captionsetup{font=footnotesize}
    \caption{Q8bot is the first open-source quadruped robot constructed with no wires and cables.}
    \label{fig:heroshot}
\end{figure}

\begin{table*}[t]
\centering
\captionsetup{justification=centering}
\caption{Comparison between existing legged robots and Q8bot}
\begin{tabular}{|l|c >{\centering\arraybackslash}m{1.5cm} >{\centering\arraybackslash}m{1.0cm} >{\centering\arraybackslash}m{1.0cm}|>{\centering\arraybackslash}m{2.2cm} >{\centering\arraybackslash}m{1.2cm} c  >{\centering\arraybackslash}m{1.1cm} >{\centering\arraybackslash}m{1.6cm}|}
\hline
    \rule{0pt}{15pt}Robot & DOF & Body \newline Length (m) & Mass (kg) & Cost \newline (USD) & Actuator Type & Mass \newline Ratio (\%) & Gear Ratio & Torque \newline (Nm) & Maximum \newline Speed (rad/s)\\[8pt]
    \hline
    Mini Cheetah \cite{MiniCheetah} &12 &0.38 &9    &4980 &Proprioceptive &44 &9:1      &17    &40\\
    Doggo \cite{Doggo}              &8  &0.42 &4.8  &3000 &Proprioceptive &45 &3:1      &4.8   &N/A\\
    OpenRoACH \cite{OpenRoACH}      &2  &0.15 &0.18 &150  &Brushed DC     &11 &50:1     &N/A   &65\\
    Pupper \cite{Pupper}            &12 &0.20 &2.1  &2000 &BLDC motor     &61 &36:1     &1.8   &60\\
    HyperDog \cite{Hyperdog}        &12 &0.30 &5    &N/A  &RC (PWM) servo       &36 &300:1    &7     &8\\
    Chen et al. \cite{Compliant}    &8  &0.12 &0.4  &N/A  &Smart (TTL) servo    &36 &288/77:1\textsuperscript{1} 
                                                                                    &0.6/0.23\textsuperscript{1} &13/48\textsuperscript{1}\\
    \textbf{Q8bot}                  &8  &\textbf{0.08} &\textbf{0.22} &\textbf{300}  &Smart (TTL) servo    &\textbf{58} &77:1     &0.23  &\textbf{48}\\
\hline
\end{tabular}
\vspace{0.2em}
    
    \raggedright
    \hspace{0.8em}\textsuperscript{1}: Two different actuator types were used.
\end{table*}

Recent quadruped robot designs have gradually converged to a size range that strikes a balance between portability and the ability to navigate obstacles in natural environments. However, there exists a motivation for smaller quadruped robots for applications such as education, swarm robotics research, and exploration in confined spaces \cite{OpenRoACH}. Such applications require robots that are smaller in size, more simply constructed, while still being robust and agile to achieve required tasks reliably. 

Various legged robots have been developed under the size of 15 cm \cite{OpenRoACH}, \cite{kim2006isprawl}. While these robots exhibit notable performance and durability, they have limited degrees of freedom and use mechanically linked legs with fixed gaits. A miniature quadruped recently developed by Chen et al. \cite{Compliant} features 8 degrees of freedom, which is typically the minimum required for agile legged locomotion. Chen et al.'s work uses a cable-driven mechanism tailored for a specific compliant mechanism, making it relatively unoptimized for generic research applications. Other low-cost quadruped platforms such as Petoi Bittle \cite{petoi_bittle} and MiniPupper \cite{mangdang_minipupper2} are available commercially at the size of around 20 cm. The Bittle robot achieves high performance using low-cost servos, but lacks the customizability of an open-source design. Conversely, MiniPupper is fully open-source but focuses on ROS education, offering limited agility in low-level locomotion tasks.

On the other hand, legged robots are inherently complex electro-mechanical systems. As a result, connecting individual components within them requires dozens of electrical wires of varying sizes and configurations. Improper wiring practices can lead to connection failures on legged robots, particularly during rapid and repetitive joint movements. This issue is especially critical for open-source platforms, where other researchers rely on limited assembly and wiring instructions to replicate the original robot.


Previous work on open-source legged robots have attempted to address this issue in various ways. Stanford Doggo \cite{Doggo}, Minitaur \cite{Minitaur}, and Chen et al.'s work \cite{Compliant} have actuators fixed to the main robot body. While this eliminates repetitive cable strain due to joint movement, the complexity of wiring remains largely unchanged. Stanford Pupper \cite{Pupper}, Doggo's successor from Stanford Student Robotics, uses printed circuit boards (PCB) as both structural support and electrical power distribution, effectively removing part of the complexities in robot wiring.

To address the limitations mentioned above, we introduce Q8bot, a compact, robust, and high-performance quadruped robot designed for easy replication. To the best of our knowledge, this is the first open-source legged robot constructed entirely without electrical wires. In the remaining sections of this paper, we will first detail how we carefully selected the actuators, microcontrollers, batteries, and other peripheral hardware to construct such a novel, wire-free robot architecture. Then, we will demonstrate the effectiveness of our approach by evaluating the robot based on common performance metrics, testing its robustness, and verifying its ease of replication through preliminary user study. Finally, we will discuss the implications of our design approach and possible future improvements of Q8bot.

\section{Hardware Design}
Q8bot weighs only 220 grams, and has a typical standing height of 7 cm, a fixed width of 7 cm, and a body length of 8 cm (defined by the distance between the midpoint of front and back linkages). The robot can be folded to fit within a 12 cm x 7 cm x 5 cm volume and can achieve 5.8 cm of chassis ground clearance. The first 4 columns of Table I provides a comparison between Q8bot and existing legged robots in terms of size, weight, and cost.

\begin{figure}[t]  
    \centering
    \setlength{\belowcaptionskip}{0pt}
    \includegraphics[width=0.48\textwidth]{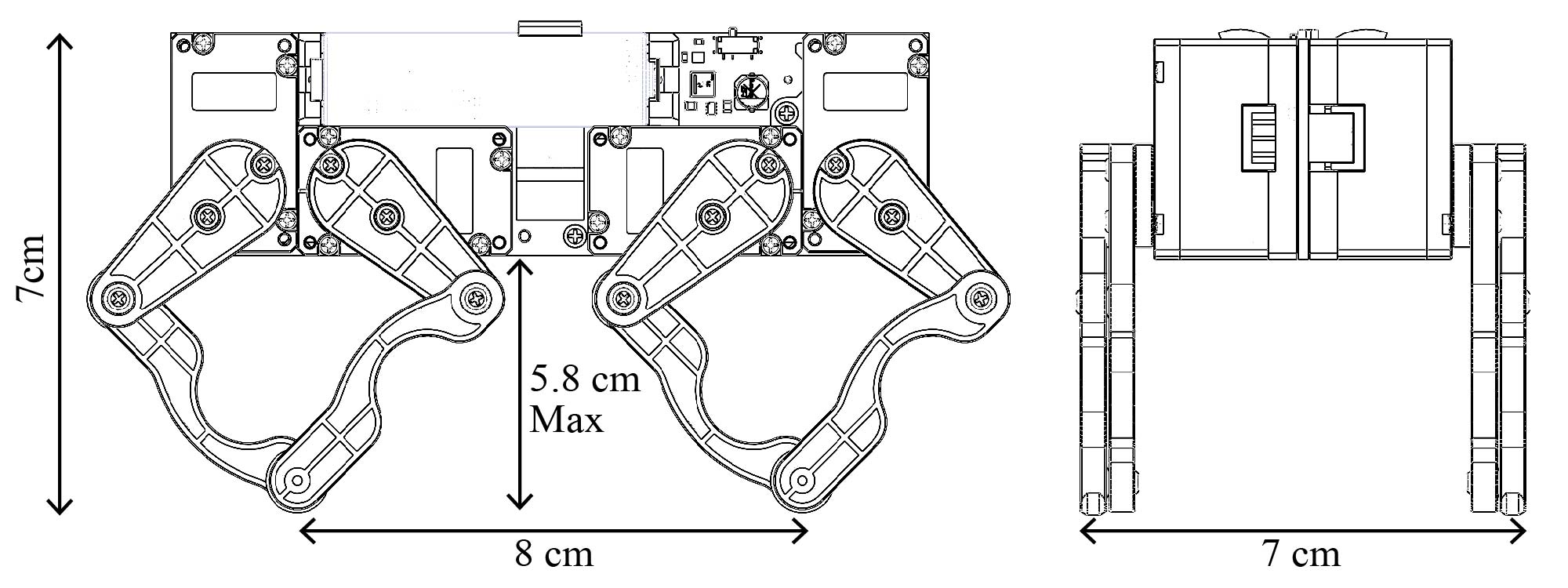}  %
    \captionsetup{font=footnotesize}
    \captionsetup{belowskip=-5pt}
    \caption{Dimension information of Q8bot.}
    \label{fig:dimensions}
\end{figure}

\begin{figure*}[t]  
    \centering
    \setlength{\belowcaptionskip}{0pt}
    \includegraphics[width=\textwidth]{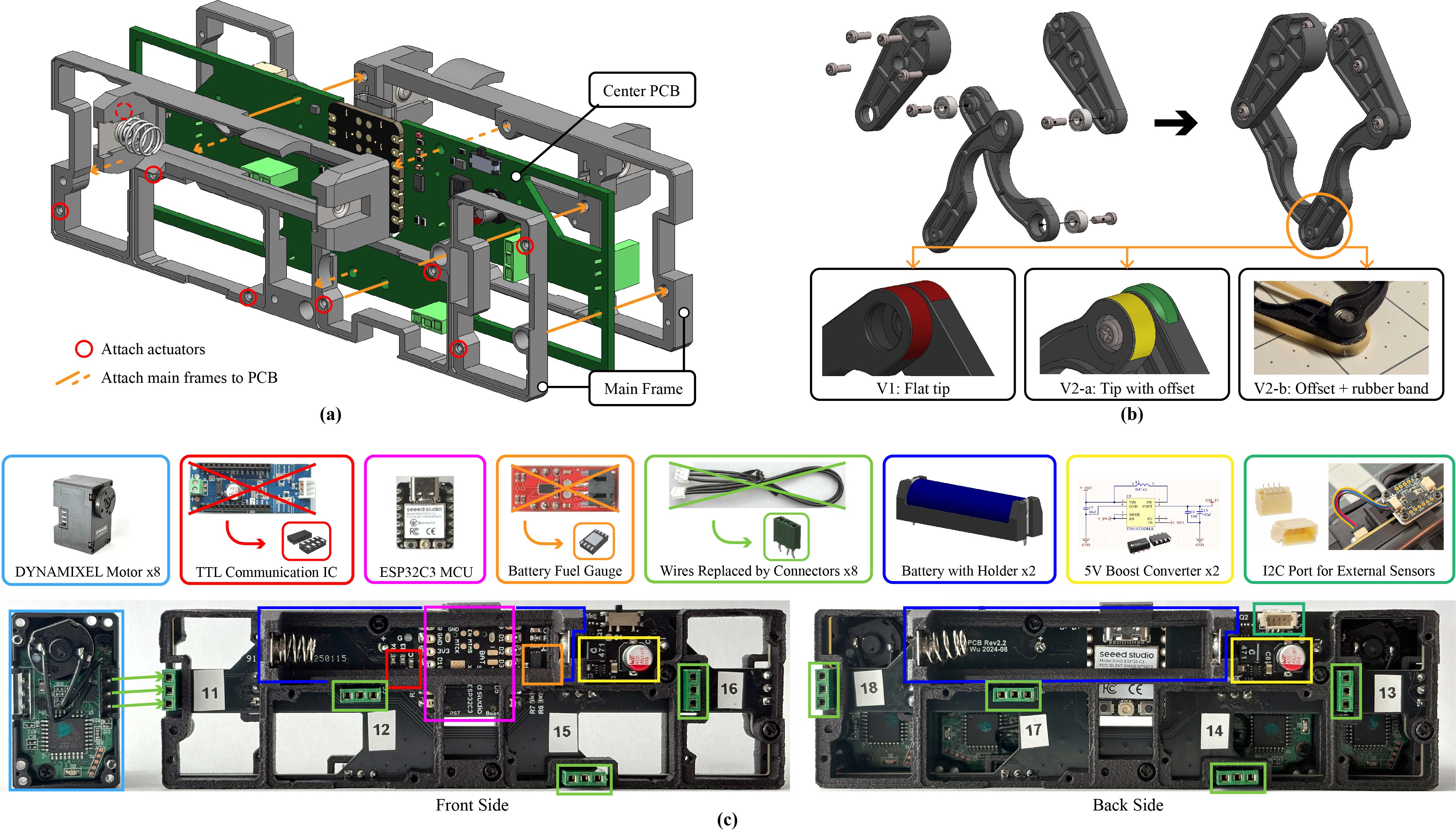}  
    \captionsetup{font=footnotesize}
    \captionsetup{belowskip=-5pt}
    \caption{Design overview of Q8bot: (a) Center spine construction, (b) leg linkage breakdown with different ground contact geometry, and (c) Direct connection of various components to eliminate wires. Eight motor TTL cables are replaced with rigid pin headers, as indicated by the green boxes in the annotation.}
    \label{fig:design}
\end{figure*}

\subsection{Actuator Selection}
Underactuated legged robots typically uses proprioceptive actuators to achieve agile motion due to their high torque density and and low reflected inertia \cite{katz2018low}. However, miniaturizing such actuators for quadruped robots as small as Q8bot remains challenging, largely because the efficiency of brushless DC (BLDC) motors decreases quickly with their size \cite{harrington2013BLDC}. As a result, most miniature or low-cost quadruped robots such as HyperDog \cite{Hyperdog} use RC servos driven by brushed DC motors, which often lacks the rotational speed and torque transparency needed to achieve agile locomotion.

Q8bot utilizes inexpensive but relatively high-performance actuator, the DYNAMIXEL XL330-M077-T from Robotis. This actuator strikes a good balance between output torque and rotational speed, and the compact form factor makes it ideal for use on miniature, dynamic legged robots. With an onboard microcontroller and absolute encoder, the XL330 can be controlled via DYNAMIXEL’s TTL protocol, allowing communication with individual motors via a shared serial bus using unique motor IDs. A comparison between the XL330 and other actuators used in various legged robots are summarized in the last 5 columns of Table I.

\subsection{Zero-Wire Approach}
To reduce electronics footprint and eliminate connecting wires, we integrated various discrete components onto a single printed circuit board (PCB) as shown in Fig.~\ref{fig:design}. This PCB is positioned vertically at the center of Q8bot, matching the cross-sectional size of the chassis, and serves as both a functional and structural element. Two identical, 3D-printed main frames are attached to the PCB from both sides via self-tapping screws to form a rigid ``spine", while also containing geometry for holding the cylindrical batteries. The latest design also features an I2C expansion port, which enables easy integration of additional sensor hardware, such as an inertia measurement unit (IMU).

The XL330 series actuators, as shown in Fig.~\ref{fig:design}c, has three-pin male connectors pointing to the back. While intended to be used with the DYNAMIXEL TTL cables, they also enable all motors to connect directly to the custom PCB using standard 2.54 mm pin headers. To secure the actuators, we removed their back plates and attached them to the main frames using their built-in screws. The resulting stack-up is a rigid, rectangular robot chassis with all electrical connections converging at the center, either through soldering or plug-and-play connectors.

\begin{figure}[t]  
    \centering
    \setlength{\belowcaptionskip}{0pt}
    \includegraphics[width=0.48\textwidth, trim=0 0 0 2cm, clip]{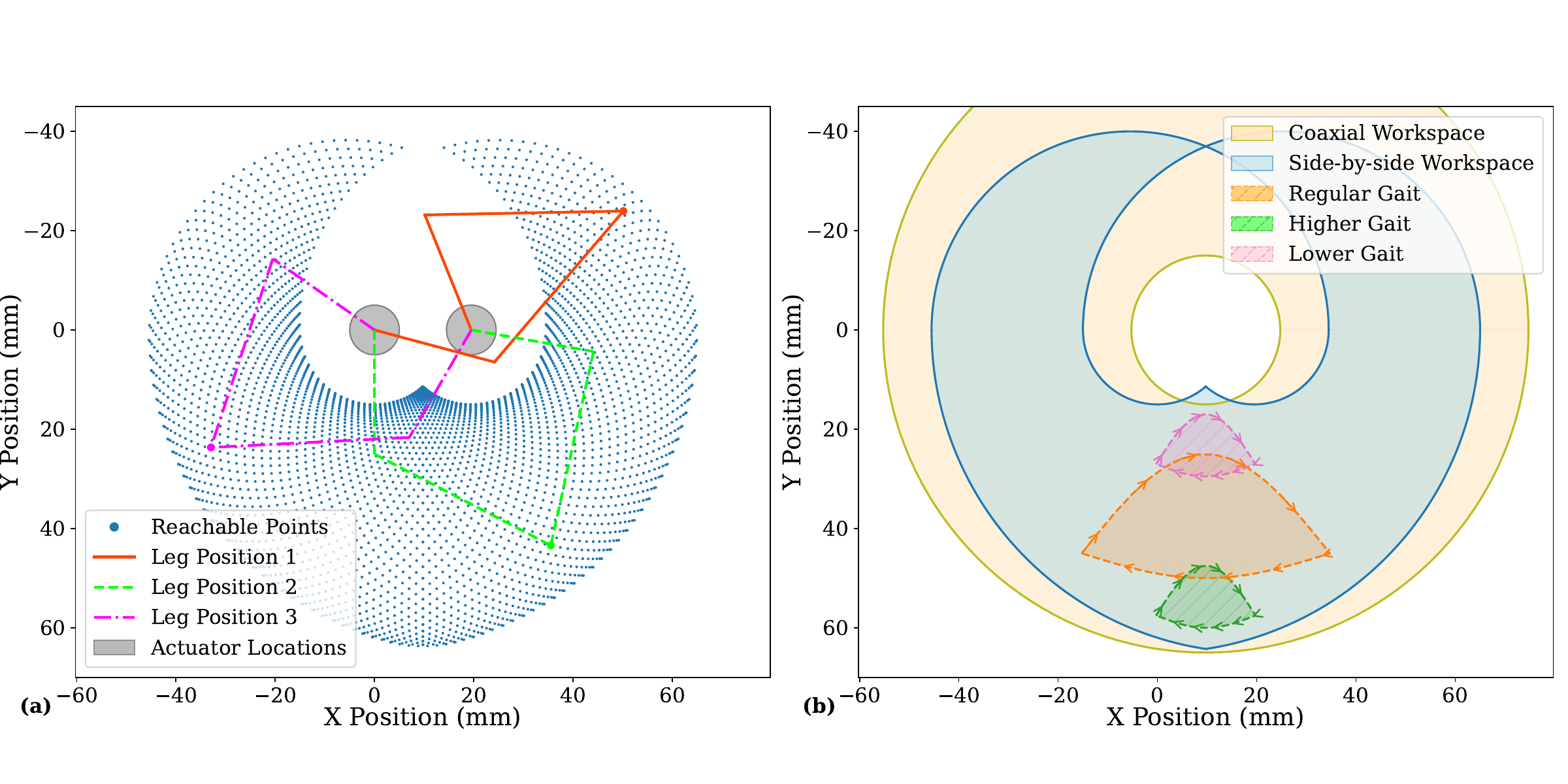}  %
    \captionsetup{font=footnotesize}
    \captionsetup{belowskip=-5pt}
    \caption{(a) Theoretical workspace of Q8bot. Manipulability increases as density decreases. (b) Workspace comparison between Q8bot's side-by-side mounting design and a coaxial design.}
    \label{fig:workspace}
\end{figure}

\subsection{Leg Geometry Design}
Q8bot has 8 degrees of freedom (DOF) and features four identical five-bar leg linkages, similar to the designs of Stanford Doggo \cite{Doggo} and Minitaur \cite{Minitaur}. The key difference is that we mounted each actuator pair side by side, rather than coaxially, which simplifies the mechanical stackup. As shown in Fig.~\ref{fig:workspace}b, a side-by-side motor placement reduces the theoretical reachable workspace but has minimal impact on the gait trajectories implemented in this study. Our leg geometry also prevents continuous rotation, which may limit the robot’s ability to perform back-flips or two-leg standing---both beyond the scope of this work. Each leg's construction consists of four 3D-printed segments, three small ball bearings, and three self-tapping screws. Four additional screws are used to attach each leg linkage to the robot chassis.

As shown in Fig.~\ref{fig:design}b, we also explored different ground contact geometry for Q8bot. We found that having a flat tip on the leg linkage leads to more stable movements in the sagittal plane but slows down turning maneuvers due to the opposite rolling direction of the two leg segments. Therefore, we introduced an offset on the outer segments to ensure the legs roll in a single direction, which noticeably improved in-place turning performance of the robot. A rubber band or similar grippy material can also be added to the tip to improve the robot's movement on inclines or under payload. We will be using legs with different contact geometry interchangeably in the following experiments, as the simple attachment method makes it easy to swap designs depending on the application.

\subsection{Software and Control}
\begin{figure}[t]  
    \centering
    \includegraphics[width=0.489\textwidth]{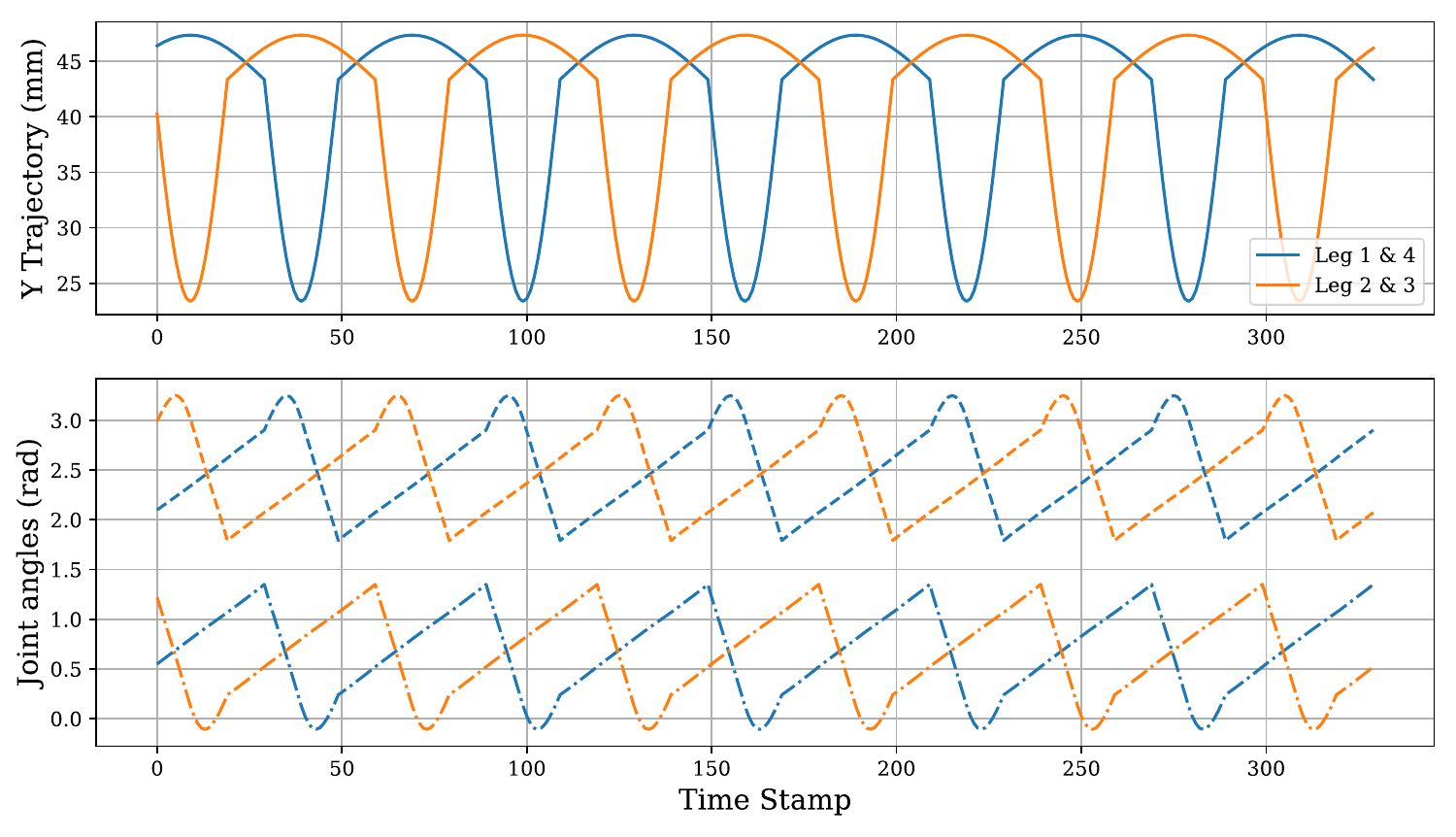}  
    \captionsetup{font=footnotesize}
    \caption{Trotting gait: generated trajectory and joint angles.}
    \label{fig:traj}
\end{figure}

At the lower level, Q8bot generates open-loop gait trajectory based on the sinusoid equation:
\[y(x) = y_0 \pm y_{\textit{range}} \sin(\pi x)\]
where \(y_0\) is the resting position and \(y_{\textit{range}}\) is the desired amplitude. Two sinusoids corresponding to the lift phase and ground phase of the leg are combined to form a repeatable loop. The individual trajectory of each of the four legs are generated by offsetting the generated array in different sequence to create various robot gaits such as walking, trotting, and bounding. Fig.~\ref{fig:traj} shows an example trotting trajectory and corresponding joint angles for diagonally paired legs.

\begin{figure}[t]  
    \centering
    \includegraphics[width=0.489\textwidth]{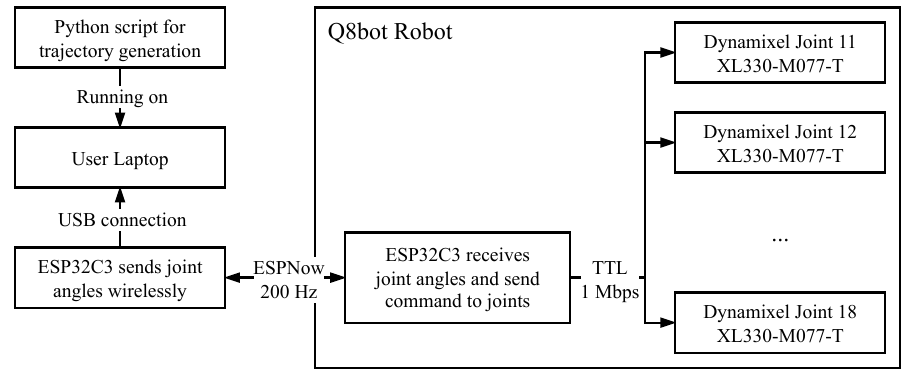}  
    \captionsetup{font=footnotesize}
    \captionsetup{belowskip=-10pt}
    \caption{Remote control framework for Q8bot.}
    \label{fig:flowchart}
\end{figure}

The software control architecture of the robot is optimized for teleoperation and gait exploration. The gait trajectories are pre-computed on a host laptop, which then sends joint commands directly to the robot at 200 Hz using the ESPNow protocol from Espressif. This enabled us to rapidly test various gaits empirically by easily modifying the underlying gait generation parameters such as sinusoid amplitude, ratio of lift phase to ground phase, and offset patterns between legs. Fig.~\ref{fig:flowchart} provides an overview of the Q8bot control framework; additional details are available in our recently accepted manuscript \cite{wu2025q8bot}.

\section{Results}

\begin{figure}[t]
    \centering
    \includegraphics[width=0.489\textwidth]{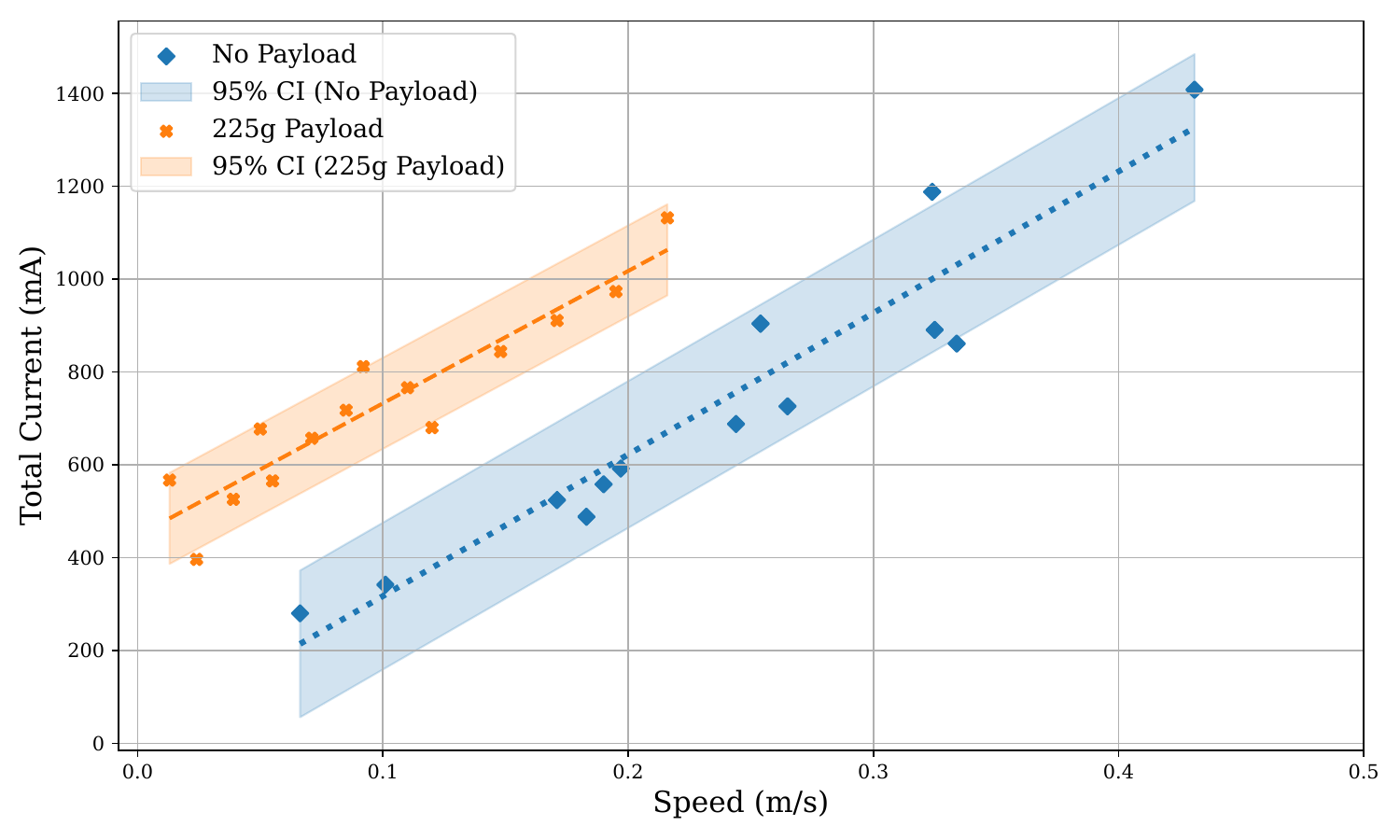}  
    \captionsetup{font=footnotesize}
    \captionsetup{belowskip=-10pt}
    \caption{Current consumption versus trotting speed. }
    \label{fig:payload}
\end{figure}

\subsection{Robot Performance}
With its unique construction method, a large portion of the robot's mass comes from the motors (Table I). As a result, Q8bot can produce high actuator speeds despite their low rated torque, enabling various tunable, dynamic gaits. To evaluate its locomotion performance, we commanded the robot to trot in a straight line under different gait parameters and measured the average current consumption of the actuators during the movement. We then used the Tracker video analysis software \cite{tracker} to extract the speed of Q8bot and graphed each data point against the corresponding current consumption in Fig.~\ref{fig:payload}. The same experiment was repeated with a payload equivalent to the robot's own weight.

The cost of transport (COT) is often used to evaluate the locomotion efficiency of legged robots \cite{Doggo}. It can be calculated with the robot's input voltage \(V\), average current consumption \(i\), mass \(m\), and steady-state velocity \(v_{ss}\) : 

\[COT = \frac{Vi}{mg v_{ss}}\]

Throughout the experiment, Q8bot achieved a lowest COT of 6.04 with a speed of 0.18 m/s. This value is relatively high compared to other existing legged robots (Table II), largely due to the low-efficiency, economic actuators used. The highest stable trotting speed was recorded at 0.43 m/s, which normalized to 5.4 body lengths per second. In a separate experiment, the robot achieved an impressive turning rate of 5 rad/s, exceeding that of many existing dynamic quadrupeds such as the MIT Mini Cheetah \cite{MiniCheetah}. 

Q8bot can perform an adequate jumping motion and reach a peak vertical jump height of 7 cm, which is measured by the change of center of gravity from the crouched position to the apex of the jump \cite{Doggo}. Although largely consistent with the single-leg experiments, the jumping performance of the standalone robot is limited by the maximum current output of the onboard boost converter. The robot also demonstrates slope climbing capability up to 20 degrees with a slow trotting gait and a low body position. Preliminary tests showed Q8bot successfully carrying a weight of 440 g during locomotion, roughly twice its own weight. Table II summarizes the Q8bot's performance in comparison with other legged robot platforms. More detailed documentation of the robot’s capabilities is available in the accompanying video submission and GitHub repository.

\begin{table*}[t]
\centering
\captionsetup{justification=centering}
\caption{Performance comparison between existing quadruped robots and Q8bot}
\begin{tabular}{|l|>{\centering\arraybackslash}m{1.8cm} >{\centering\arraybackslash}m{1.6cm} >{\centering\arraybackslash}m{1.6cm} >{\centering\arraybackslash}m{1.6cm} >{\centering\arraybackslash}m{1.6cm} >{\centering\arraybackslash}m{1.6cm} >{\centering\arraybackslash}m{1.2cm}|}
\hline
    \rule{0pt}{15pt}Robot & Ground Velocity (m/s) & Normalized Speed (\(s^{-1}\)) & Turning Rate (rad/s) & Max Payload (kg) & Normalized Payload & Normalized Workload & COT \\[8pt]
    \hline
    Mini Cheetah \cite{MiniCheetah} & 2.45 & 6.45 & 4.6 & 9\textsuperscript{*} 
                                                               & 1.0  & 6.45  & N/A\\
    Doggo \cite{Doggo}              & 0.8  & 2.14 & N/A & N/A  & N/A  & N/A   & 3.2\\
    OpenRoACH \cite{OpenRoACH}      & 0.34 & 2.27 & N/A & 0.2  & 1.04 & 2.36  & 7.7\textsuperscript{*}\\
    Pupper \cite{Pupper}            & 0.8  & 4.0  & 2.5 & N/A  & N/A  & N/A   & 5.5\textsuperscript{*}\\
    HyperDog \cite{Hyperdog}        & 0.3\textsuperscript{*}  
                                           & 1.0\textsuperscript{*}  
                                                  & N/A & 2    & 0.4  & N/A   & 6.0\textsuperscript{*}\\
    Chen et al. \cite{Compliant}    & 0.52 & 4.4  & N/A & N/A  & N/A  & N/A   & 3.2\\
    \textbf{Q8bot}                  & 0.43 &\textbf{5.38} &\textbf{5.0} & 0.22 & \textbf{2.0}  & \textbf{10.76} & 6.04\\
\hline
\end{tabular}
\vspace{0.2em}
    
    \raggedright
    \hspace{3.5em}\textsuperscript{*}: Estimated values based on published data. Normalized Work Capacity = normalized
speed × normalized payload.
\end{table*}

\subsection{Platform Robustness}
\begin{figure}[t]
    \centering
    \includegraphics[width=0.489\textwidth]{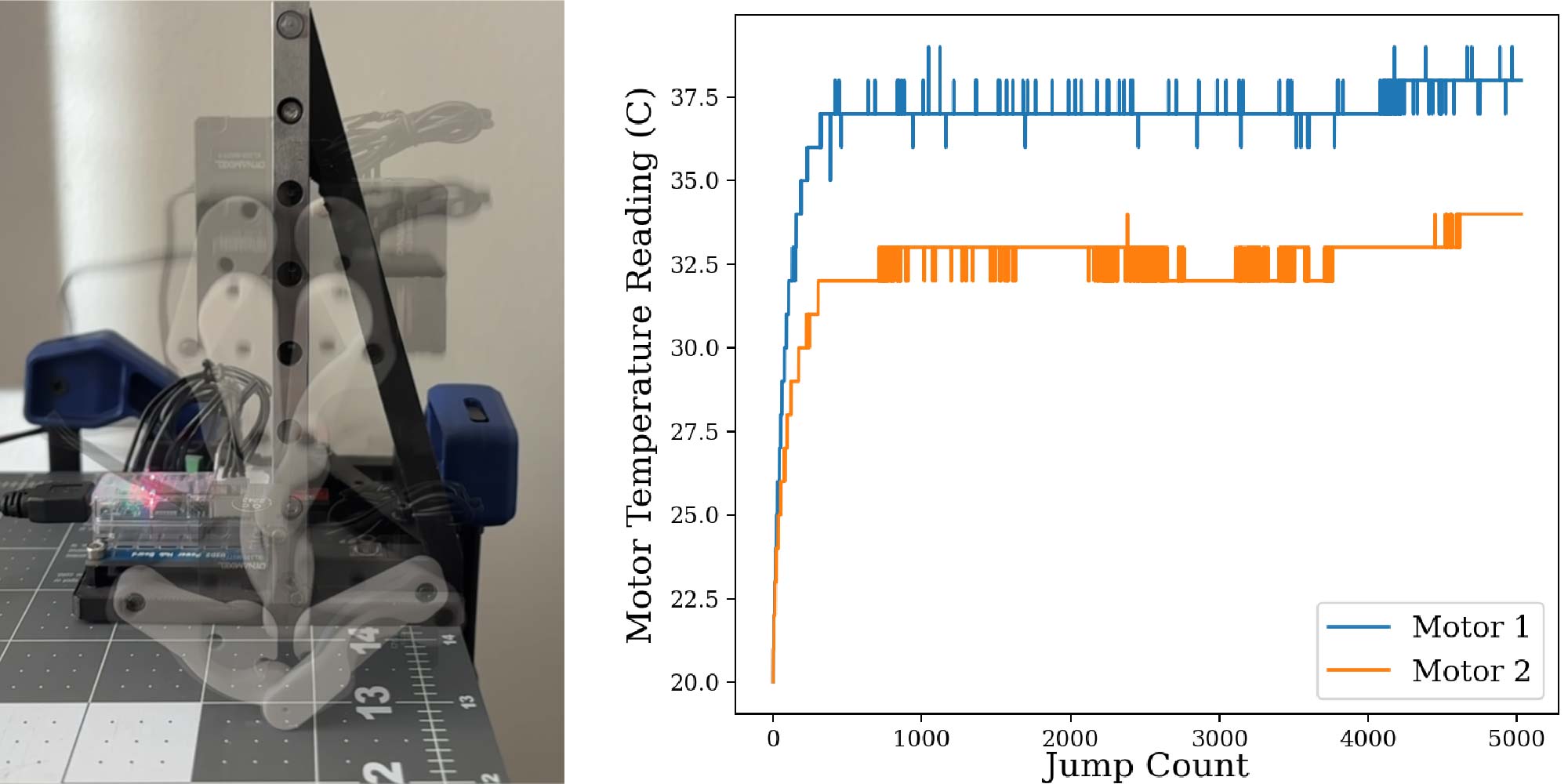}  
    \captionsetup{font=footnotesize}
    \caption{Jumping durability test. Left: single-leg setup performing jumping motion. Right: motor temperature versus increasing jump count.}
    \label{fig:jumpCombined}
\end{figure}

To evaluate the durability of Q8bot's design, we performed a jumping burn-in test using a single-leg setup shown in Fig.~\ref{fig:jumpCombined}. The setup performed more than 5000 jumps across a span of 5 hours with the actuator temperature stabilizing at 37.5 degrees measured from the actuators' built-in sensors. The burn-in test was repeated multiple times, and no damage was observed to the motors or gearboxes. 

An endurance test was conducted on the standalone Q8bot, where we teleoperated the robot to run back and forth in a 12-meter-long hallway while continuously monitoring the battery level using the onboard fuel gauge. During the experiment, the robot used 60 \% of its combined battery capacity of 2000 mAh and covered approximately 650 meters in 1 hour. This corresponds to an average speed of 0.18 m/s, or a normalized speed of 1.5 body lengths per second. Since this calculation did not account for the time spent turning around at the end of the hallway segment, the true average speed is likely higher than the calculated value.

\begin{table}[t]
\centering
\captionsetup{justification=centering}
\caption{Q8bot vertical drop test}
\begin{tabular}{|l|>{\centering\arraybackslash}m{1cm} >{\centering\arraybackslash}m{1cm} >{\centering\arraybackslash}m{1cm} >{\centering\arraybackslash}m{1cm} >{\centering\arraybackslash}m{1cm}|}
\hline
\rule{0pt}{10pt}Height                     & Top    &Side    &Bottom                   &Front                    &Tumble \\ [4pt]
\hline
\rule{0pt}{10pt}25 cm\textsuperscript{*}   &Success &Success &Success                  &Success                  &Success\\ [4pt]
\rule{0pt}{10pt}50 cm\textsuperscript{*}   &Success &Success &Success                  &Success                  &Success\\ [4pt]
\rule{0pt}{10pt}75 cm\textsuperscript{*}   &Success &Success &\textbf{Gearbox failure} &\textbf{Gearbox failure} &Success\\ [4pt]
\rule{0pt}{10pt}75 cm\textsuperscript{**}  &Success &Success &Success                  &Success                  &Success\\ [4pt]
\rule{0pt}{10pt}100 cm\textsuperscript{**} &Success &Success &\textbf{Gearbox failure} &\textbf{Gearbox failure} &Success\\ [4pt]
\hline
\end{tabular}
\vspace{0.2em}

\raggedright
\hspace{0.8em}\textsuperscript{*}: With motor torque enabled.\\
\hspace{0.8em}\textsuperscript{**}: With motor torque disabled.
\end{table}

Q8bot can be operated on a desk due to its small form factor. Therefore, we also conducted a series of drop tests to verify the mechanical robustness of the platform in case of accidental falls from an average desk height. We dropped the robot from increasing height above hardwood floors at various orientations and observed its structural integrity. As summarized in Table III, failures in individual motor gearboxes began at 75 cm when motor torque is enabled, and at 100 cm when torque is disabled. The failures occurred when the bottom or front of the robot made initial contact with the ground, causing the actuators to absorb most of the impact through the leg linkages. In comparison, the robot chassis showed greater impact resistance, indicating the effectiveness of the optimized mechanical stack-up. 

Between trials, we were able to easily swap out the failed motor due to the design simplicity of the robot chassis, limiting each repair cycle in under 2 minutes. The failed motors were repairable by replacing a spur gear inside the gearbox, which proved to be primary failure point across multiple motor units in our experiments.

\subsection{Ease of Replication}
To validate the reproducibility of Q8bot as an open-source platform, we asked 11 participants with varying levels of engineering skills to assemble the robot following building instructions in a digital PDF format. Participants include high school students, graphic design graduate students, undergraduate and graduate students majoring in mechanical engineering, as well as faculty from the UCLA Department of Mechanical Engineering. One step, which involves soldering four through-hole joints for the battery clips, was omitted from the trials due to its irreversibility, and a fixed time of 5 minutes was added to all participants' assembly time. Overall, the average assembly time for the robot was 55.3 minutes with a standard deviation of 7.3 minutes. The robot remains fully functional and structurally intact after more than 15 cycles of assembly and disassembly. 

\section{Discussion}

\subsection{Limitation and Future Directions in Control}
Since the current control strategy is limited to open-loop using pre-computed trajectories and gaits, the robot cannot actively respond to external disturbances such as changes in terrain geometry. We plan to address this shortcoming by designing a state estimator based on an extended Kalman filter (EKF) that fuses IMU data and joint encoder feedback \cite{ALPHRED}. The estimator would enable the use of model-based controllers and support the deployment of reinforcement learning policies on the physical robot. Implementing these closed-loop strategies likely requires upgrading the onboard microcontroller unit to a single-board computer such as the Raspberry Pi, which will also enable the use of more capable middleware like the Robot Operating System (ROS).

Without proprioception, Q8bot cannot reliably estimate ground contact using actuator current feedback alone. To improve the robot's performance, a simple binary foot switch can be added to each leg, enabling us to replace the heuristic gait scheduler with a finite state machine that detects unexpected contact events and performs appropriate actions \cite{Bledt}.

\subsection{Design Choices}
To reduce cost and complexity, most 3D-printed parts on Q8bot are identical copies placed on the opposite sides, minimizing the number of unique components and lowering manufacturing cost. The latest design only requires fasteners included with the DYNAMIXEL XL-330 motors, with twelve small ball bearings as the only additional hardware. Bearing fits were also fine-tuned in the 3D-printed parts to avoid the need for additional tools when installing the bearings while ensuring a secure fit during robot operation. 

For all experiments conducted in this paper, we chose to use sensors within the DYNAMIXEL actuators to record various data. While the sampling rate and accuracy could be improved by using external sensors in a custom setup, we believe this demonstrates that Q8bot is a convenient system for young researchers and students to conduct experiments without the need for additional hardware.

\subsection{Open Source and Outreach}
The Q8bot GitHub repository provides all hardware and software resources needed for replication. While custom electronics are essential to achieving Q8bot's compact form factor, recent advances in PCB manufacturing make it easy to fabricate the robot's custom board through online vendors. All structural components are designed to be 3D-printed, with two replication options: Option 1 is optimized for multi-jet fusion (MJF) technology, which researchers can order through e-commerce, and option 2 is designed for standard hobbyist FDM printers for rapid in-house prototyping.

The compact form factor and all-in-one packaging of Q8bot makes it practical to carry multiple copies to conduct research and education in remote locations. At the time of this writing, we have brought Q8bot to local high schools and college makerspaces for outreach and workshops in the hopes of inspiring young students to pursue higher education in the fields of robotics and engineering. 

\section{Conclusion}
In this paper, we introduced a wire-free, miniature quadruped robot capable of traversing various terrains and performing dynamic movements based on open-loop control. We presented the design methodology of the robot in detail, and moved on to demonstrating that Q8bot matches---and in some cases exceeds---performance of other quadruped robots based on common metrics, with the added benefits of improved robustness and replicability. This robot is ideal for applications that require efficient management and maintenance of identical legged robots, such as hands-on engineering education and swarm robotics research. Its exceptional robustness also enables rapid algorithm deployment and efficient training data collection on real hardware. 

In addition to exploring closed-loop and learning-based control methods mentioned above, we also plan to optimize Q8bot's hardware system. This includes an upgrade to the power management electronics to support more torque-demanding locomotion, as well as an improved leg mounting interface to allow faster swapping between leg configurations based on the application. By making Q8bot an open-source project, we hope to provide an accessible yet capable platform for advancing education and research in legged robotics.

\section*{Acknowledgment}

We thank A. Zhu, Y. Tanaka, Q. Wang, and M. Zhu for feedback on preliminary drafts, and Helen Xie for her assistance in creating various figures in this paper. We also thank members of UCLA Robotics and Mechanisms Lab who participated in the user assembly study of the platform.

\bibliographystyle{ieeetr}
\bibliography{main}
\end{document}